\documentclass{article}

\PassOptionsToPackage{numbers, compress}{natbib}

\usepackage[preprint]{neurips_2026}

\usepackage[utf8]{inputenc} 
\usepackage[T1]{fontenc}    
\usepackage{hyperref}       
\usepackage{url}            
\usepackage{booktabs}       
\usepackage{amsfonts}       
\usepackage{nicefrac}       
\usepackage{microtype}      
\usepackage{xcolor}         
\usepackage{multirow}

\newcommand{\COMMENT}[1]{}

\usepackage{amsmath,amssymb}
\usepackage{amsthm}
\usepackage{graphicx}

\newtheorem{theorem}{Theorem}[section]
\title{Moment kernels: a simple and scalable approach for equivariance to rotations and reflections in deep convolutional networks}

\author{%
Siqi Fang$^{1}$ \and
Zachary Schlamowitz$^{1}$ \and
Andrew Bennecke$^{1}$ \and
Daniel J. Tward$^{1,2}$ \\
$^{1}$Department of Computational Medicine, University of California, Los Angeles \\
$^{2}$Department of Neurology, University of California, Los Angeles \\
\texttt{\{sqfang, zschlamowitz, benaqui624, dtward\}@ucla.edu}
}

\begin{document}

\maketitle

\begin{abstract}
Translation equivariance is a central reason convolutional neural networks have been successful in computer vision. Other symmetries, such as rotations and reflections, are similarly important in fields such as biomedical image analysis, but equivariant methods for these symmetries remain less widely adopted, especially in 3D. Existing approaches often rely on group convolutions, harmonic bases, irreducible representations, or specialized libraries, which can obscure the explicit form of admissible kernels for practitioners. 
We introduce moment kernels, a simple Cartesian parameterization of convolution kernels equivariant to orthogonal transformations, $O(d)$, between tensor-valued feature fields. We prove that every such $O(d)$-equivariant kernel can be represented as a sum of radial functions of $|x|$ multiplied by products of coordinate components $x^i$ and Kronecker deltas. This gives a complete, dimension-agnostic kernel family complementary to harmonic-basis approaches and implementable using standard convolution modules.
We implement a discrete version of moment-kernel networks and evaluate on biomedical tasks with different transformation laws: invariant 2D image classification and equivariant 3D affine-transform regression for brain MRI. Across these tasks, moment kernels improve worst-case orientation consistency and remain trainable in 3D, while avoiding the orientation-channel expansion required by group convolutions, which reaches 48 orientations for 90-degree rotations and reflections in 3D. The resulting models provide exact consistency under grid-preserving rotations and reflections, and remain practical for standard CNN workflows.
\end{abstract}

\section{Introduction}
While several technologies have worked together to bring about the success of deep learning in computer vision, one major early advancement was the convolutional neural network (CNN)\citep{lecun1989handwritten}. Unlike multi-layer perceptrons, CNNs leverage the principle of translation equivariance: translating the input image results in an equivalent translation of the output. 
Today, CNNs have been adopted for nearly all applications in computer vision. Despite the success of translation equivariance, rotation- and reflection-equivariant methods remain uncommon in biomedical image-analysis pipelines, especially in 3D. In many such settings, image orientation is arbitrary, so different views of the same sample should produce consistent predictions. Equivariant models provide this consistency by construction, which is particularly valuable in medical applications where reliability and trust are essential.

Several approaches exploit rotation and reflection symmetries. Group convolutions \citep{cohen2016group} augment feature maps with group-valued orientation channels, yielding exact equivariance for the chosen group but increasing memory and computation with the group size; even 90-degree rotations and reflections require 8 orientations in 2D and 48 in 3D. Steerable-kernel methods \citep{freeman1991design,weiler20183d,bekkers2018roto} avoid explicitly storing every transformed feature map, but their general formulations are typically expressed through harmonic bases, irreducible representations, or specialized libraries. These methods are powerful, but their mathematical and implementation overhead can be a barrier in some biomedical workflows.

A complementary line of work treats feature channels as components of geometric objects such as scalars, vectors, matrices, or higher-order tensors \citep{thomas2018tensor,cesa2022a,lang2020wigner}. These quantities transform directly under image rotations and reflections; for example, a vector-valued feature's components rotate with the input image. This provides an interpretable transformation law (relative to coefficients of harmonic expansions) and avoids combinatorial growth in number of orientation channels. Our goal is to give a complete Cartesian characterization for the important case of orthogonal symmetries, using only geometric quantities familiar in basic image analysis: coordinates, Kronecker deltas, scalars, vectors, and matrices (and for completeness, other tensors).

Here, we develop a complete, dimension-agnostic characterization of convolution kernels equivariant to orthogonal transformations between tensor-valued feature fields. For a rank-$r$ tensor-valued kernel, every admissible term is a radial function of $|x|$ multiplied by products of coordinate components $x^i$ and Kronecker deltas $\delta^{ij}$, with indices arranged according to a finite set of pairings. We call these moment kernels. The characterization is complete, yet remains in Cartesian coordinates and can be derived using elementary linear algebra. No irreducible representations, Clebsch-Gordan coefficients, or specialized libraries are required. The same formulas extend to any dimension $d\geq2$ without re-derivation, unlike (for example) the progression from Fourier series in 2D to spherical harmonics in 3D. In the continuous setting they describe O($d$)-equivariant kernels (O($d$) is the group of orthogonal transformations in $d$ dimensions, i.e. rotations and reflections). On discrete voxel grids, exact equivariance is retained for grid-preserving rotations and reflections, with other angles handled approximately through interpolation.

We evaluate the approach on biomedical tasks requiring different output transformation laws: invariant classification of 2D medical images and equivariant 3D affine-transform regression for brain MRI. These experiments test both the reliability benefit of exact worst-case consistency, where moment kernels achieve zero worst-case drop by construction across DermaMNIST and BloodMNIST, and the practical trainability of the architecture in 3D, where moment kernels make end-to-end O(3) affine regression trainable with default hyperparameters.

\section{Moment kernels}
We first characterize how geometric quantities transform under rotations and reflections (§\,\ref{sec:transform}), then derive what constraints equivariance places on convolution kernels (§\,\ref{sec:kernels}), and finally identify all solutions: the moment kernels (§\,\ref{sec:moment_kernels}). All results hold for any dimension $d\geq 2$, so the same formulas extend from 2D to 3D without re-derivation.

\subsection{Equivariant transformation laws for geometric functions}
\label{sec:transform}
Let $R$ be an orthonormal transform (rotation/reflection $\textstyle R^{-1}\!\!=\!\!R^T$) acting on a function. The function's components transform according to its geometric character: scalar, vector, matrix, or general tensor.

\paragraph{Definitions.}
A map $\Phi$ between geometric function spaces is \emph{equivariant} with respect to $O(d)$ if
\begin{align}
\label{eq:equivariance_def}
    \Phi(R \cdot f) = R \cdot \Phi(f) \quad \text{for all inputs } f \text{ and all } R \in O(d),
\end{align}
where ``$\cdot$'' denotes the appropriate group action on each function space.
As a special case, a map is \emph{invariant} if the group acts trivially on the output, i.e., $\Phi(R \cdot f) = \Phi(f)$.
For the classification tasks we consider, the output is a predicted class label, which should be unaffected by image orientation (invariance).
For registration tasks, the output (a matrix-valued field) transforms non-trivially with $R$, so the full equivariance condition applies.

\paragraph{Field transformations.}
For scalar, vector, matrix, and rank-$r$ tensor fields, the action of $R$ is:
%
\begin{align}
&\text{($r\!\!=\!\!0$): } [R\!\cdot\! f](x) \!=\! f(R^{-1}x), \  
\text{(} r\!\!=\!\!1\text{): } [R\!\cdot\! v](x) \!=\! R\,v(R^{-1}x),\ 
\text{(} r\!\!=\!\!2\text{): } [R\!\cdot\! M](x) \!=\! R\,M(R^{-1}x)R^T  \notag \\
&\text{(arbitrary rank } r \text{): } [R\cdot T]^{i_1,\ldots,i_r}(x) = R^{i_1}{}_{i'_1}\cdots R^{i_r}{}_{i'_r}\,T^{i'_1,\ldots,i'_r}(R^{-1}x)
\label{eq:tensor}
\end{align}
Each law rotates the domain point $x$ and the geometric components independently. The rank-$r$ form specializes to scalar, vector, and matrix cases above. Repeated indices are summed (Einstein convention). Since $R$ is orthogonal, transformations of covariant and contravariant components coincide, and we use raised-index position throughout for our kernels.  Such notation is common in applied treatments of differential geometry, such as fluid dynamics \cite{cohen2007fluid}\footnote{%
The rotation matrix $R^i{}_j$ is written with a lowered second index to emphasize $R$'s action on tangent vectors; the transpose is $(R^T)^i{}_j = R^j{}_i$. Readers may substitute the fully raised form $R^{ij}$ throughout without loss of correctness.}

\subsection{Characterizing equivariant convolution kernels}
\label{sec:kernels}
Applying the equivariance condition \eqref{eq:equivariance_def} for $\Phi$ to a convolution constrains the kernel. We show in Appendix~\ref{app:law} that this constraint takes a clean form: an equivariant rank-$r$ kernel must transform exactly as a rank-$r$ tensor field,
\begin{align}
\label{eq:kernel_general}
    k^{i_1,\ldots,i_r}(Rx) = R^{i_1}{}_{i'_1}\cdots R^{i_r}{}_{i'_r}\, k^{i'_1,\ldots,i'_r}(x) \ ,
\end{align}
and it can map a rank-$i$ input (for any $i \in \{0,\ldots,r\}$) to a rank-$(r-i)$ output by contracting over its last $i$ indices.  For this reason, using superscript indices exclusively simplifies notation. The four cases relevant to scalar and vector features are the specializations to $r=0,1,1,2$, each obtained from \eqref{eq:kernel_general} by the standard substitution-on-the-integrand argument (full derivation in Appendix~\ref{app:law}):
\begin{align}
    \begin{array}{llll}
    k(Rx) = k(x) & \text{(scalar$\to$scalar),} & k(Rx)=Rk(x) & \text{(scalar$\to$vector),}\\
    k(Rx) = k(x)\,R^T & \text{(vector$\to$scalar), } & k(Rx) = R\,k(x)\,R^T & \text{(vector$\to$vector)}
    \end{array}
    \label{eq:example_kernel_transformations}
\end{align}

\subsection{Characterization of moment kernels}
\label{sec:moment_kernels}
We now solve the master equation \eqref{eq:kernel_general}. Our main theoretical result is that the solution space has a clean combinatorial structure, which we call the \emph{moment kernel family}. A convolution with such a kernel computes weighted spatial moments of the input, just as the $r$th moment of a function involves integration against $x^r$.

\begin{theorem}[Completeness of moment kernels; proved in Appendix~\ref{app:moment_all}]
\label{thm:completeness}
Let $d \geq 2$ and $r \geq 0$. Every kernel $k$ satisfying the master equation \eqref{eq:kernel_general} takes the form
\begin{align}
\label{eq:moment_kernel}
\textstyle k^{i_1,\ldots,i_r}(x) = \sum_\sigma f_\sigma(|x|)\;\delta^{i_{\sigma_{1,a}}i_{\sigma_{1,b}}}\cdots\delta^{i_{\sigma_{s,a}}i_{\sigma_{s,b}}}\; x^{i_{\sigma^C_1}}\cdots x^{i_{\sigma^C_{r-2s}}} \ ,
\end{align}
where the sum is over all \emph{signatures} $\sigma$, sets of non-overlapping index pairs drawn from $\{1,\ldots,r\}$, and each $f_\sigma:\mathbb{R}_{\geq 0}\to\mathbb{R}$ is a learnable radial function. 
Here we describe a given by $\sigma = \{\{\sigma_{1,a},\sigma_{1,b}\},\ldots,\{\sigma_{s,a},\sigma_{s,b}\}\}$ for $s = |\sigma|$ total index pairs. We denote its complement $\sigma^C = \{1,\ldots,r\}\setminus \cup \sigma = \{\sigma^C_{1},\ldots,\sigma^C_{r-2s}\}$ for $|\sigma^C|=r-2s$ unpaired indices. Each index pair $\{a,b\} \in \sigma $ contributes a Kronecker delta $\delta^{i_a i_b}$.  Each remaining unpaired index $c \in \sigma^C$ is filled by a position component $x^{i_c}$.
\end{theorem}

The characterization is complete: no $O(d)$-equivariant kernel lies outside this family. The parameterization uses only two geometric primitives (Kronecker deltas and position components) multiplied by rotation-invariant radial functions and extends to arbitrary dimension $d \geq 2$ without modification.

\paragraph{Invariant building blocks.} To see where Theorem~\ref{thm:completeness} comes from, note three quantities that transform correctly under $R$:
(i) a radial function $f(|x|)$, invariant since $|Rx|=|x|$;
(ii) a position component $x^i$, a rank-1 tensor since $(Rx)^i = R^i{}_j x^j$;
(iii) the Kronecker delta $\delta^{ij}$, an invariant rank-2 tensor since $R^i{}_k R^j{}_l \delta^{kl} = (RR^T)^{ij} = \delta^{ij}$.
Any product with $r$ unpaired indices automatically satisfies \eqref{eq:kernel_general}. The theorem's content is that these building blocks exhaust the space of solutions.

\paragraph{Low-rank cases.} For low ranks Theorem~\ref{thm:completeness} gives compact closed forms. Examples are in Table~\ref{tab:example_low_rank}. 
\begin{table}[h]
\caption{\label{tab:example_low_rank} Example low rank moment kernels}
\small
\begin{tabular}{clll}
\toprule
$r$ & Example mapping(s) & Moment kernel form & \# signatures \\
\midrule
0 & scalar$\to$scalar                            & $k(x) = f_{\varnothing}(|x|)$                                       & 1 \\
1 & scalar$\to$vector, vector$\to$scalar       & $k^i(x) = f_{\varnothing}(|x|)\,x^i$                                & 1 \\
2 & vector$\to$vector, scalar$\to$matrix       & $k^{ij}(x) = f_{\{\{1,2\}\}}(|x|)\,\delta^{ij} + f_{\varnothing}(|x|)\,x^i x^j$ & 2 \\
3 & vector$\to$matrix, \ldots                    & one $x^ix^jx^k$ term $+$ three $\delta^{\cdot\cdot}x^\cdot$ terms & 4 \\
4 & matrix$\to$matrix                            & $x^ix^jx^kx^l$ $+$ six $\delta^{\cdot\cdot} x^{\cdot} x^{\cdot}$ terms $+$ three $\delta^{\cdot\cdot}\delta^{\cdot\cdot}$ terms & 10 \\
\bottomrule
\end{tabular}
\end{table}%
\begin{figure}[h]
\centering
\includegraphics[width=0.9\linewidth]{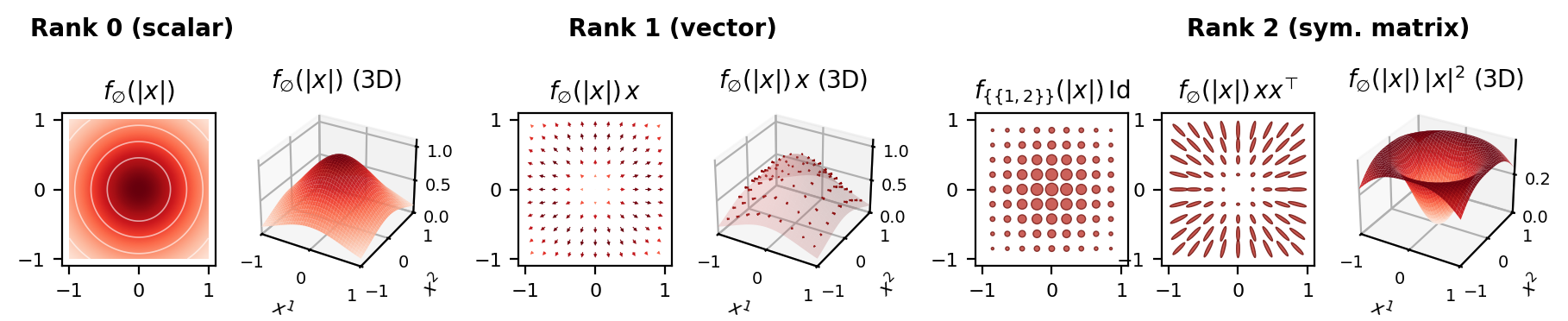}
\vspace{-0.6em}
\caption{\small Basis elements of moment kernels for $r=0,1,2$. Rank 2 visualized as ellipsoids.}
\label{fig:kernels}
\vspace{-1em}
\end{figure}%
The rank-0 kernel is radially symmetric, so it cannot encode orientation (e.g., edge detection requires $r\geq 1$). This is a potential advantage of our method over the SRE-Conv framework \cite{du2024sreconv} which only uses kernels of this form. The rank-2 form, a linear combination of isotropic and dyadic terms, previously appeared in shape-analysis applications \citep{micheli2013matrix}. In general, the number of signatures grows as
$N(r) = \sum_{s=0}^{\lfloor r/2\rfloor}\binom{r}{2s}\,(2s-1)!!$
where $s$ counts the number of index pairs and $(2s-1)!!$ (where the double factorial multiplies odd integers up to $2s-1$) gives the number of pairings of $2s$ indices.

\subsection{Implementation}
\label{sec:implementation}
We define a radial function along one axis with a fixed number of samples (default 3), and resample it into a hypercube (here $3\!\times\!3$, or $3\!\times\!3\!\times\!3$) for convolution using linear interpolation.  Interpolation weights are precomputed and executed as matrix multiplication.  When a convolution module is initialized, all signatures for a given rank $r$ are enumerated, and one radial function is randomly initialized per signature.  The kernel is constructed as an array with $r\times d$ axes, and reshaped using lexicographic ordering to give the correct input and output dimensions ($i\times d$ and $(r-i)\times d$), where $i$ is the rank of the input tensor field. Kernels of different ranks are stacked into a single matrix-valued kernel whose blocks map between tensor fields of different rank, allowing the full equivariant linear operation on all tensor-valued features to be executed as one standard convolution call.
In the discrete setting ($d=2$), the integer grid is invariant only under the dihedral group $D_4$, the 8-element group of 90$^\circ$ rotations and reflections.  Consequently, both moment kernel networks and ESCNN \citep{cesa2022a} are exactly equivariant under $D_4$ and approximately equivariant at other angles. This is a shared limitation of any method that applies a fixed discrete kernel to a square integer grid, not a property of the moment kernel parameterization specifically.

\subsection{Architecture details}
\label{sec:architecture_details}
\paragraph{Bias.}
Vectors and any tensor fields of odd rank must have a bias of 0.  We use a standard bias for scalars and a constant multiple of identity for matrices. We do not explore bias for other even rank tensors in this work, as they are not used in the experiments.

\paragraph{Pointwise nonlinearities.}
We apply a rectified linear unit to the log magnitude of a tensor at every point (adding a small constant $\epsilon$ inside the square root of the magnitude to avoid division by zero), and exponentiate the result. As such, any tensor with magnitude less than one is rescaled to have magnitude 1. Other choices, such as a gated nonlinearity \cite{cesa2022a}, have been explored elsewhere. Since this operation depends only on the magnitude of each tensor, which is invariant under rotation and reflection, the direction is unchanged and equivariance is preserved.

\paragraph{Batch normalization.}
We continue the simple approach to nonlinearities and simply apply a standard batch normalization layer \citep{ioffe2015batch} to the log magnitude. We multiply by a scalar factor such that the variance is 1 after taking the exponential (i.e., lognormal assumption). We note that this scales tensors so their magnitude is 1 on average, which centers them at the elbow of our nonlinearity above. Scaling by a scalar factor preserves the tensor direction and therefore preserves equivariance. 

\paragraph{Downsampling by two.}
If an image is even size in a given dimension, we downsample by averaging neighboring pixels.  If odd, we downsample by subsampling neighboring pixels.  Downsampling by subsampling only (for example applying a ``stride'' in convolution layers) will not be equivariant if the dimension is even, because the first pixel is never skipped, and the last pixel always is.  Taking a reflection would exchange the roles of first and last, giving a different result.

\section{Experimental Design}
We include two experiments to illustrate the benefits of invariance and equivariance in biomedical applications and to quantify the performance of our architectures relative to standard CNN baselines and existing equivariant methods.  We provide implementation code in the supplementary material.

\subsection{Image Classification}
\label{sec:image_classification}
We evaluate on DermaMNIST, a 7-class skin-lesion classification benchmark from MedMNIST \citep{yang2023medmnist} (CC BY-NC 4.0 license). Not all classification tasks are expected to be invariant to rotation: chest X-rays have a canonical orientation and left versus right carry clinical meaning. Skin-lesion appearance is genuinely rotation-invariant, and published accuracy on this benchmark is well below ceiling effects. We follow a standard protocol: multiclass AUC is computed one-versus-rest and the checkpoint with the best validation AUC selects the test-set prediction \citep{yang2023medmnist}.

We use a simple 8-layer architecture. Each layer consists of a convolution, batch normalization, and pointwise nonlinearity; channels double and spatial resolution halves after every second layer. Our moment kernel model uses all kernel ranks up to 4 (scalar, vector, and matrix channels), outputs one invariant scalar per class, and has 174k parameters. We define \emph{worst-case accuracy} (WC accuracy) as the fraction of test images correctly classified under all 8 orientations (4 rotations $\times$ 2 reflections). A single incorrect prediction in any orientation counts as a failure for that image.

We compare to three baselines.
(i)~\textbf{Standard CNN}: same 8-layer architecture without equivariance constraints, 200k parameters.
(ii)~\textbf{ESCNN} \citep{cesa2022a}: 185k parameters, exact $O(2)$-equivariance via irreducible representations using \texttt{flipRot2dOnR2} with \texttt{GNormBatchNorm} and \texttt{NormNonLinearity}.
(iii)~\textbf{SRE-Conv kernel}: the spatially-averaged filter technique of \citet{du2024sreconv} substituted into our 8-layer backbone (210k parameters). The architecture downsamples four times via stride-2 average pooling, so we resize DermaMNIST to $32\times32$ to preserve $D_4$-equivariance on the integer grid (full integration details in Appendix~\ref{sec:sre-conv-ablations}).

All models were trained for 1,000 epochs with Adam \citep{kingma2014adam} on an NVIDIA GeForce RTX 4090, except ESCNN under continuous augmentation which required 3,000 epochs to converge (indicated by $^\dagger$). We evaluate three augmentation conditions: none, discrete (random 90$^\circ$ rotations and reflections), and continuous (uniform random angle, zero padding).

\subsection{Image registration}\label{sec:registration}
Image registration is the task of applying a spatial transformation to one image (a target) to bring it into correspondence with another (an atlas). Among other applications, this is used in brain image processing to make comparisons across populations of images. Different spatial transformations have been used in both classical \cite{klein2009evaluation} and deep learning settings \cite{yang2017quicksilver, balakrishnan2019voxelmorph, hoffmann2021synthmorph}. We use a 12-parameter affine transformation in 3D. Equivariant networks have been used for feature matching in 2D bioimages \cite{wang2024rotir} and natural images \cite{bokman2022case}, and for 3D medical-image registration via steerable-kernel encoders coupled to non-equivariant decoders or closed-form solvers \cite{rezvani2026, billot2024equitrack}. To our knowledge, our work is the first to use equivariance throughout the network for end-to-end 3D affine regression.

\paragraph{Task and data.}
We use 3,745 brain MRI volumes from the Alzheimer's Disease Neuroimaging Initiative (ADNI), aligned to an ICBM atlas \cite{fonov2009unbiased} using the pipeline of \cite{puglisi2024enhancing} and resampled to $80\times80\times80$ voxels. Data are split 80/10/10 by subject (no subject appears in more than one set); the test set comprises 412 volumes. To protect privacy, the skull and face of each participant are removed.  Secondary analysis of anonymized imaging data is generally considered not human subjects research.  During training we apply (using linear interpolation) a known random affine transformation to each volume and train the network to predict it by minimizing MSE on affine matrix elements. Applying such transforms occasionally moves imaging data outside the field of view, making the task approximately but not exactly equivariant.
When rotating or reflecting a volume, each column of the predicted affine matrix transforms like a 3D vector, which our equivariant architecture enforces by construction. We evaluate regression quality (per-element MSE on $\hat A$, worst-orientation MSE, and inter-orientation prediction inconsistency). Downstream alignment-quality metrics such as Dice on tissue masks, target registration error on landmarks, or post-warp image similarity are complementary to these and are left to future work.

\paragraph{Methods compared.}
We compare six model configurations spanning matched, medium, and large parameter regimes: (i)~{Moment kernel} (proposed, 174k): the architecture introduced in %
§\ref{sec:image_classification} but with $d=3$ and 4 output vector channels, exactly $\mathrm{O}(3)$-equivariant.
(ii)~\textbf{Standard CNN} (187k, parameter-matched): the same backbone with standard 3D convolutions, non-equivariant. (iii)~\textbf{SynthMorph-style encoder} \cite{hoffmann2021synthmorph} at default 5.4M and parameter-matched 175k variants: a non-equivariant encoder-plus-dense-head architecture from the registration literature, trained from scratch under our supervised affine-regression objective (Appendix~\ref{sec:synthmorph-arch}). (iv)~\textbf{ESCNN} \cite{cesa2022a} at two configurations representing the two paradigms in published 3D ESCNN work: $\mathrm{SO}(3)$ (matching \citet{rezvani2026}'s encoder; 435k) and $\mathrm{O}(3)$ (with architectural reflection equivariance; 417k). We report the no-batch-normalization variant for both groups: removing batch normalization was the only setting we found that allowed the $\mathrm{SO}(3)$ model to train, and we report the $\mathrm{O}(3)$ variant under the same setting to test whether this fix transfers to reflection-equivariant ESCNN. Eight ESCNN variants in total were tested across group, irrep mix, nonlinearity, batch normalization, and capacity (Appendix~\ref{sec:escnn-3d-attempts}).

\paragraph{Training and evaluation.}
All methods are trained for 200 epochs with Adam (lr $ = 10^{-4}$, batch size 16). We evaluate on the held-out test set under 48 axis-aligned orientations, the full octahedral group with reflections. 
For each test volume $v$ and orientation $R$, we measure the per-element squared error of the matrix-valued prediction, averaged across the 12 affine elements:
$\mathrm{SE}(v, R) = \tfrac{1}{12} \sum_{i,j} (\hat{A}_{ij} - (RA)_{ij})^2$.
From this we report three metrics: (i)~\textbf{MSE}: mean of $\mathrm{SE}(v, R)$ over test volumes and orientations, the per-element mean squared error, in the same units as the training loss. (ii)~\textbf{WC-MSE}: per-volume maximum of $\mathrm{SE}(v, R)$ over orientations, then averaged over test volumes (worst case orientation per-element MSE, only reported in the appendices). (iii)~\textbf{ED2}: mean squared distance between predictions on the same volume across pairs of orientations, averaged over the 12 elements of the predicted affine. Predictions are compared in canonical frame as $R^{-1} \hat{A}_R$, embedded into $4 \times 4$ homogeneous form for the inverse. The bottom row $[0,0,0,1]$ is identical across orientations by construction and contributes 0 to the squared distance. ED2$=0$ means perfectly consistent predictions regardless of input orientation. In practice, even an exactly equivariant model has a small but non-zero ED2 floor in this protocol.  This is due to input images being generated through interpolation, with occasional clipping of image data outside the field of view. (iv)~\textbf{WC-MSE rot}: 
per-volume maximum of $SE(v,R)$ over orientations $R$ with no reflection, then averaged over test volumes (worst case rotation-only per-element MSE).

Metrics are split by rotation vs.\ reflection for fair comparison against architectures with rotation-only equivariance: ``MSE rot'' and ``WC-MSE rot'' restrict the mean and the per-volume maximum, respectively, to the 24 rotation orientations. Accidentally reflecting neuroimaging data is a surprisingly common pitfall \cite{glen2020beware}, which can be mitigated through equivariance to reflections.

We run three experiments. Experiment 1 (the main comparison with results shown in Table \ref{tab:registration}, six methods) trains all methods with $\mathrm{SO}(3)$ augmentation. When evaluated on $\mathrm{O}(3)$ transformed images, we expect moment kernels and $\mathrm{O}(3)$ ESCNN to perform well, and other models to perform poorly. Experiments 2 and 3 are restricted to moment kernel and standard CNN. Experiment 2 trains both with no rotation augmentation, testing whether equivariance is structural or must be learned. Experiment 3 trains the standard CNN with full $\mathrm{O}(3)$ augmentation, testing whether data augmentation can substitute for architectural equivariance.

\section{Results}
\subsection{Image Classification results}

Moment kernels (174k parameters) achieve WC drop $= 0\%$ under all three augmentation conditions  (Table~\ref{tab:classification}) and across all five random seeds, confirming that exact equivariance holds regardless of training augmentation or initialization. 
The standard CNN at 200k parameters achieves comparable  accuracy at 76.1\% with continuous augmentation but substantially lower WC accuracy (55.0\% without augmentation and 62--64\% with augmentation), consistent with lack of equivariance.  

We compare to two equivariant or approximately-equivariant baselines. ESCNN \citep{cesa2022a} (185k parameters, discrete augmentation) achieves 74.0\% accuracy and 74.0\% WC accuracy (WC drop $= 0\%$), confirming that exact $D_4$ equivariance is preserved under our evaluation protocol. However, it is $2.7\times$ slower per epoch (10.5\,s vs.\ 3.9\,s, measured back-to-back on the same GPU). The three architectures respond differently to augmentation. For moment kernels, all three augmentation conditions produce statistically indistinguishable results, as predicted by exact equivariance. ESCNN, by contrast, is hurt by continuous augmentation: training was extended to $3{,}000$ epochs to allow convergence and still reached lower accuracy ($70.3\%$ vs.\ $74.0\%$ with discrete augmentation), bringing total training cost to ${\approx}8\times$ that of moment kernels ($3\times$ more epochs at $2.7\times$ slower per epoch). Both moment kernels and ESCNN are exactly equivariant on the integer grid, so this asymmetry under continuous augmentation likely reflects representation-specific factors in the ESCNN design rather than a general consequence of equivariance. We leave identification of a precise mechanism to future work. The SRE-Conv kernel \citep{du2024sreconv} substituted into our 8-layer backbone (210k parameters) is also exactly $D_4$-equivariant on the integer grid (Table~\ref{tab:classification}, WC drop $= 0\%$ across all three augmentation conditions) when trained at 32$\times$32 input (Table~\ref{tab:classification} footnote $\ddagger$), but achieves lower accuracy (70.8\%-71.7\%) than moment kernels (75.4\%-76.6\%) at the same parameter scale.  MedMNIST publishes accuracy and AUC for seven non-equivariant methods on their website (\url{https://medmnist.com/}) and our method achieves the best AUC and second-best accuracy (best was 76.8\% for ``Google AutoML Vision'') among these. 

A tensor-rank ablation isolating the contribution of vector and matrix channels (rank 0 vs.\ 2 vs.\ 4) is reported in Appendix~\ref{sec:rank-ablation}. Equivariance is preserved at every rank, and accuracy increases monotonically from 67.6\% (scalars only) to 76.6\% (full rank-4).

\begin{table}[t]
  \centering
  \caption{ Image classification on DermaMNIST test set. WC accuracy = fraction of images
  correctly classified in all 8 orientations (4 rotations $\times$ 2 reflections). \textbf{Bold} indicates  best result in each column.}
  \label{tab:classification}
  \small
  \begin{tabular}{llrrrrr}
    \toprule
    Model & Aug & Params & Acc & WC Acc & AUC \\
    \midrule
    Moment kernels (ours)            & none       & 174k & 75.4\%          & 75.3\%          & 0.925          \\
    Moment kernels (ours)            & discrete   & 174k & \textbf{76.6\%} & \textbf{76.6\%} & 0.932          \\
    Moment kernels (ours)            & continuous & 174k & \textbf{76.6\%} & \textbf{76.6\%} & \textbf{0.933} \\
    ESCNN \citep{cesa2022a}          & none       & 185k & 72.1\%          & 72.1\%          & 0.874 \\
    ESCNN \citep{cesa2022a}          & discrete   & 185k & 74.0\%          & 74.0\%          & 0.889 \\
    ESCNN \citep{cesa2022a}$^\dagger$ & continuous & 185k & 70.3\%          & 70.3\%          & 0.864 \\
    SRE-Conv \citep{du2024sreconv}$^{\ast,\ddagger}$       & none       & 210k & 71.7\%          & 71.7\%          & 0.892          \\
    SRE-Conv \citep{du2024sreconv}$^{\ddagger}$       & discrete   & 210k & 71.1\%          & 71.1\%          & 0.897          \\
    SRE-Conv \citep{du2024sreconv}$^{\ddagger}$       & continuous & 210k & 70.8\%          & 70.8\%          & 0.899          \\
    Standard CNN                     & none       & 200k & 72.0\%          & 55.0\%          & 0.898 \\
    Standard CNN                     & discrete   & 200k & 75.2\%          & 62.1\%          & 0.920 \\
    Standard CNN                     & continuous & 200k & 76.1\%          & 63.9\%          & 0.931 \\
    \bottomrule
  \end{tabular}
  \par\smallskip
  {\footnotesize \quad $^{\dagger}$\,ESCNN under continuous augmentation required 3{,}000 epochs to converge (vs.\ 1{,}000 for all other rows). \\
  $^{\ast}$\,The SRE-Conv kernel is integrated into our 8-layer 210k backbone for parameter-controlled comparison. The published SRE-ResNet18 architecture (3.99M parameters) and alternative integration variants are evaluated separately in Appendix~\ref{sec:sre-conv-ablations}. \\
  $^{\ddagger}$\,SRE-Conv uses 32$\times$32 resize $+$ \texttt{AvgPool2d(2,2)}, matching the published recipe; 
  other variants reported in Appendix~\ref{sec:sre-conv-ablations}.
  }
\end{table}


\paragraph{Variability.} To assess robustness across initialization for our method, we repeated  analysis with 4 additional random seeds. The standard deviation of accuracy ($=$ worst-case accuracy) and AUC across these 5 runs is $1.10\%/0.0045$ (no aug), $1.11\%/0.0093$ (discrete), and $0.69\%/0.0051$ (continuous).

\paragraph{Off-axis rotations.}
The WC accuracy in Table~\ref{tab:classification} measures consistency across 8 $D_4$-aligned orientations. To probe behavior at other angles, we sweep the test set through 36 evenly-spaced rotations (from $0^\circ$ to $350^\circ$, replicate-pad resampling) and compute maximum accuracy drop relative to $0^\circ$ per seed, averaged across 5 random seeds.  We also examined the effect of interpolation (from a radial function of one axis to a $3\times 3$ kernel). Linear interpolation shows max drops of $1.23 \pm 0.70\%$ (no aug), $0.96 \pm 0.41\%$ (discrete aug), and $0.73 \pm 0.69\%$ (continuous aug). Cubic interpolation gives a comparable $1.10 \pm 0.56\%$ (no aug), indicating that  rotation behavior is robust to the choice of interpolation. Combined with the WC drop $= 0\%$ guarantee at the 8 $D_4$-aligned orientations, this confirms that moment kernels remain approximately rotation-consistent at arbitrary angles.
  

\paragraph{Generalization to a second dataset.}
We replicate the same protocol on BloodMNIST, an 8-class peripheral blood cell classification benchmark from MedMNIST \citep{yang2023medmnist}. Moment kernels again achieve the highest WC accuracy (96.2\%) with zero WC drop without any rotation augmentation. ESCNN and SRE-Conv also have zero WC drop but lower WC accuracy, while the standard CNN loses 3.7--17.1\% under the worst-case protocol. Full results and discussion are deferred to Appendix~\ref{sec:bloodmnist}.

\subsection{Image registration results}


\begin{table}[t]
\centering
\caption{3D affine registration on ADNI test set (412 volumes $\times$ 48 orientations). All methods trained with $\mathrm{SO}(3)$ augmentation. All metrics are per-element averages over the 12 elements of the predicted affine, in the same units as the training loss. \textbf{Bold} indicates the best result in each column.}
\label{tab:registration}
\small
\begin{tabular}{lrccccc}
\hline
Method & Params & MSE (all) & MSE rot & MSE refl & ED2 & WC-MSE rot \\
\hline
Moment kernels (ours) & 174k & $\mathbf{0.0126}$ & $\mathbf{0.0126}$ & $\mathbf{0.0126}$ & $\mathbf{0.0048}$ & $\mathbf{0.0178}$ \\
Standard CNN                  & 187k & 0.196 & 0.0345 & 0.358 & 0.182 & 0.0768 \\
SynthMorph-style (default)    & 5{,}399k & 0.209 & 0.0341 & 0.384 & 0.207 & 0.0801 \\
SynthMorph-style (matched)    & 175k & 0.231 & 0.188 & 0.273 & 0.122 & 0.358 \\
ESCNN $\mathrm{SO}(3)$ (no-BN)& 435k & 0.201 & 0.0573 & 0.345 & 0.193 & 0.130 \\
ESCNN $\mathrm{O}(3)$ (no-BN) & 417k & 1.893 & 1.895 & 1.890 & 1.147 & 3.093 \\
\hline
\end{tabular}
\end{table}

\paragraph{Moment kernels uniquely combine accuracy and orientation consistency.}
Moment kernels are the only architecture we compared that simultaneously achieves accurate affine regression (test MSE 0.0126) and low orientation inconsistency (ED2 0.0048) (Table~\ref{tab:registration}). The other five methods exhibit one of two failure modes. (i) Insufficient symmetry coverage: the standard CNN, SynthMorph-style default, and ESCNN $\mathrm{SO}(3)$ variant fit rotations competitively (MSE rot 0.034-0.057) but error increases by an order of magnitude under reflections (MSE refl 0.34-0.38, a factor of $6$-$11\times$), as we expected for this experimental design,
and predictions vary across orientations (ED2 0.18-0.21). None of these three provide $\mathrm{O}(3)$ symmetry: the standard CNN and SynthMorph-style encoder are non-equivariant; ESCNN $\mathrm{SO}(3)$ is architecturally rotation-equivariant but lacks reflection equivariance. (ii) Failure to fit the training distribution: the SynthMorph-style matched variant (MSE rot 0.19) and the ESCNN $\mathrm{O}(3)$ variant (MSE rot 1.90) do not even adequately fit rotations, so accuracy is  $15$-$150\times$ worse than  moment kernels before reflections enter the picture. The lower performance of O(3) ESCNN in this configuration was not expected, so we explored many different architecture variants as described in Appendix~\ref{sec:escnn-3d-attempts}.
The ESCNN configurations using batch normalization reach a similar plateau, with validation MSE in the range 1.6-2.0 across the architectural axes we explored (Appendix~\ref{sec:escnn-3d-attempts}). The combination of both exact $\mathrm{O}(3)$ equivariance and end-to-end trainability with default settings is what moment kernels uniquely provide. Because the standard CNN has more parameters than the moment kernel, its performance gap cannot be attributed to capacity.

\paragraph{Encoder architecture alone does not close the gap.}
The SynthMorph-style encoder  \citep{hoffmann2021synthmorph}, trained under our supervised affine-regression objective, does not close the gap: the default 5.4M-parameter variant matches the standard CNN at 187k (test MSE 0.209 vs.\ 0.196) and the parameter-matched 175k variant reaches MSE 0.231 ($18\times$ worse than the moment kernels). Both variants exhibit the same rotation-vs-reflection gap as the standard CNN (MSE rot 0.034-0.19, MSE refl 0.27-0.38). The conclusion is restricted to the encoder design under our supervised objective, not to the released SynthMorph as a registration system. Full architectural and training details are in Appendix~\ref{sec:synthmorph-arch}.

\paragraph{End-to-end 3D $\mathrm{O}(3)$ regression with ESCNN does not match moment kernels under default settings.}
Our ESCNN comparison is restricted to end-to-end regression of the affine transformation matrix from a single input volume with default hyperparameters. Across the eight variants explored (Appendix~\ref{sec:escnn-3d-attempts}), every batch-normalized configuration plateaus at validation MSE 1.6-2.0 for both \texttt{IIDBatchNorm3d} and the equivariance-preserving \texttt{GNormBatchNorm}, consistent with \citet{billot2024equitrack}'s vanishing-gradient observation. Removing batch normalization allows training at the cost of either reflection equivariance (our $\mathrm{SO}(3)$ variant) or accuracy ($\mathrm{O}(3)$ variant, val\ MSE 1.79). Published 3D ESCNN works \citep{rezvani2026, billot2024equitrack} couple ESCNN to a separate inference component (closed-form solver, non-equivariant decoder), which we did not reproduce. We make the narrow claim that moment kernels make end-to-end 3D $\mathrm{O}(3)$ affine regression trainable with default settings where end-to-end ESCNN regression does not.

Two additional experiments confirm that the gap is structural: training both methods without rotation augmentation and giving the standard CNN full $\mathrm{O}(3)$ augmentation against the $\mathrm{SO}(3)$-trained moment kernel. The moment kernel handles unseen rotations consistently while the standard CNN collapses (ED2 0.097 vs.\ 1.77), and $\mathrm{O}(3)$ augmentation more than halves the standard-CNN MSE but leaves it $6.5\times$ behind the moment kernel. Full results are in Appendix~\ref{sec:registration-aug}.

\section{Discussion}
\paragraph{Summary.}
In this work we characterized a new kind of convolution kernel equivariant to rotations and reflections, termed a ``moment kernel,''  which operates on images of geometric features (scalars, vectors, matrices, or other tensors).  In image classification, we demonstrated competitive  accuracy and showed particularly strong results for ``worst case accuracy.'' These accuracy and worst-case-consistency results are stable across 5 random seeds and extend to off-$D_4$ rotation angles where moment kernels remain approximately rotation-consistent. This worst-case measure is particularly relevant to biomedical usage: different orientations of the same image are expected to receive the same classification label. Providing a worst-case scenario lower bound fortifies trust in ML methods. In 3D affine image registration, the moment kernel is the only architecture in our six-method comparison that simultaneously achieves accurate regression and orientation-consistent predictions, with $15\times$ lower MSE and $38\times$ lower inter-orientation inconsistency than a parameter-matched standard CNN over the full 48-element octahedral group. Three experiments demonstrate that this advantage is structural: it persists without rotation training and is not closed by data augmentation alone. A SynthMorph-style encoder \cite{hoffmann2021synthmorph} trained under the same supervised affine-regression objective reaches only the standard-CNN level on this task even at $31\times$ the parameter count and ESCNN \cite{cesa2022a} configurations trained as end-to-end $\mathrm{O}(3)$ affine regressors with default settings either fail to train (consistent with \citet{billot2024equitrack}'s vanishing-gradient observation in 3D) or, after non-default modifications, train but lose architectural reflection equivariance without recovering moment-kernel accuracy.

\paragraph{Limitations.}
In this work we tried to make our networks simple, using standard modules where possible.  This also means there are many features that have not yet been optimized.
Moment kernels have higher per-epoch cost than a standard CNN in the 2D classification setting (3.9\,s vs.\ 1.4\,s/epoch), though they are $2.7\times$ faster than ESCNN (3.9\,s vs.\ 10.5\,s/epoch).  In our 3D registration setting, both standard CNNs and moment kernels train at $\approx44$-45\,s/epoch. One opportunity for further speedup is to exploit sparsity: a moment kernel mapping matrices to matrices is about 41\% sparse and one mapping vectors to vectors is about 37\% sparse. Exploiting this structure could reduce the 2D overhead significantly.
Other areas lacking optimization in our approach include parameter initialization, choice of nonlinearities, and normalization strategies.
Finally, our networks are exactly equivariant to 90$^\circ$ rotations and reflections, but for rotations by other angles equivariance is only approximate. We investigated cubic radial interpolation as an alternative to linear interpolation and found it produces statistically indistinguishable results across 5 seeds. A more thorough study of interpolation and sampling strategies and their interactions with kernel size remains open; for example, \cite{du2024sreconv} used $5\times 5$ or $9\times 9$ kernels to improve rotation invariance. In the registration setting, we compared end-to-end equivariant regressors but did not test the encoder-plus-separate-solver configurations of \citet{rezvani2026, billot2024equitrack}, which would form a complementary line of work.

\paragraph{Broader impact.}
A motivating goal of this work was to present equivariant CNNs in a form approachable by biomedical researchers without specialized background in differential geometry or group theory. By leveraging the known geometry of the input data, we reduce the construction of rotationally equivariant kernels to standard linear algebra and calculus. We provide several examples of rotational equivariance's utility in biomedical applications, emphasizing the value of expanding convolution kernels beyond translational equivariance. Moreover, the provably equivariant functional forms we present serve to bolster trust in ML among non-experts. Our methods can guarantee that they give consistent answers when data is viewed in different orientations and there are cases where this may be more valuable than state-of-the-art accuracy.  We began exploring this issue by considering ``worst case performance'' but this also presents an opportunity to study how users of these tools may value accuracy versus reliability.  Our proof of the completeness of the ``moment kernel'' form (Appendix \ref{app:moment_all}) further codifies the growing field of equivariance in deep learning. 
We note that the approach of ESCNN achieves the same goal using representation theory. 
For the same symmetry group and the same Cartesian tensor-field representations, the span of moment-kernels corresponds to the span of equivariant kernels constructed by representation-theoretic methods such as ESCNN, but expressed directly in Cartesian coordinates rather than harmonic bases. We believe that parameterizing kernels by tensor rank, rather than by maximum harmonic frequency, may be more intuitive for non-expert practitioners.
Finally, while we emphasized simplicity in this work, we view our tools as a complement to (not a replacement for) existing tools, which have their own unique advantages and disadvantages. 

\begin{ack}
This work was supported by NIH grants R01 NS121761 and U01 AG073804.  We gratefully acknowledge discussions with Mario Micheli regarding kernels that are linear maps from vectors to vectors for shape analysis.
Data collection and sharing for the Alzheimer's Disease Neuroimaging Initiative (ADNI) is funded by the National Institute on Aging (National Institutes of Health Grant U19AG024904). The grantee organization is the Northern California Institute for Research and Education. In the past, ADNI has also received funding from the National Institute of Biomedical Imaging and Bioengineering, the Canadian Institutes of Health Research, and private sector contributions through the Foundation for the National Institutes of Health (FNIH) including generous contributions from the following: AbbVie, Alzheimer's Association; Alzheimer's Drug Discovery Foundation; Araclon Biotech; BioClinica, Inc.; Biogen; Bristol-Myers Squibb Company; CereSpir, Inc.; Cogstate; Eisai Inc.; Elan Pharmaceuticals, Inc.; Eli Lilly and Company; EuroImmun; F. Hoffmann-La Roche Ltd and its affiliated company Genentech, Inc.; Fujirebio; GE Healthcare; IXICO Ltd.; Janssen Alzheimer Immunotherapy Research \& Development, LLC.; Johnson \& Johnson Pharmaceutical Research \& Development LLC.; Lumosity; Lundbeck; Merck \& Co., Inc.; Meso Scale Diagnostics, LLC.; NeuroRx Research; Neurotrack Technologies; Novartis Pharmaceuticals Corporation; Pfizer Inc.; Piramal Imaging; Servier; Takeda Pharmaceutical Company; and Transition Therapeutics.
\end{ack}

\newpage
\bibliography{references}

\newpage 
\appendix

\section{Proof of transformation law for tensor kernels}
\label{app:law}

We derive the transformation law on a convolution kernel required for the
convolutional map $\Phi_k$ to satisfy the equivariance condition
\[
    \Phi(R\cdot f) = R\cdot \Phi(f)
    \qquad \text{for all } f \text{ and } R\in O(d).
\]
Let $f$ be a rank-$N$ input tensor field and let $g=\Phi(f)$ be a
rank-$M$ output tensor field. In indices, the input and output actions are
\begin{align}
    [R\cdot f]^{i'_1,\ldots,i'_N}(x)
    &=
    R^{i'_1}{}_{j'_1}\cdots R^{i'_N}{}_{j'_N}
    f^{j'_1,\ldots,j'_N}(R^{-1}x), \\
    [R\cdot g]^{i_1,\ldots,i_M}(x)
    &=
    R^{i_1}{}_{j_1}\cdots R^{i_M}{}_{j_M}
    g^{j_1,\ldots,j_M}(R^{-1}x).
\end{align}
We show that equivariance of $\Phi$ is equivalent to the kernel law
\begin{align}
    k^{k_1,\ldots,k_M,j'_1,\ldots,j'_N}(y)
    =
    (R^T)^{k_1}{}_{i_1}\cdots (R^T)^{k_M}{}_{i_M}
    R^{i'_1}{}_{j'_1}\cdots R^{i'_N}{}_{j'_N}
    k^{i_1,\ldots,i_M,i'_1,\ldots,i'_N}(Ry).
\end{align}
The $M$ output indices acquire factors of $R^T$ because we solve for the
unrotated output components, while the $N$ input indices acquire factors of
$R$ from the transformed input field.

\emph{Notation.} We use the same mixed-index convention as the main text throughout both appendices: $R^i{}_j$ denotes the $(i,j)$ entry of $R$ with the output index $i$ raised and the input index $j$ lowered, so that $(R^i{}_j)(x^j) = (Rx)^i$ is the standard up-down contraction.  The transpose satisfies $(R^T)^i{}_j = R^j{}_i$; it is written explicitly as $R^T$ when it appears.

A convolution kernel is a rank $M+N$ tensor-valued function that sums over the last $N$ indices.
\begin{align}
\label{eq:match1}
    g^{i_1,\ldots,i_M}(x) = \int k^{i_1,\ldots,i_M,i'_1,\ldots,i'_N}(x-x')f^{i'_1,\ldots,i'_N}(x')dx' \ .
\end{align}
As justified in the main text, we use only raised indices on our tensor fields, and paired indices are still summed over in this notation.

Following equation \eqref{eq:tensor}, equivariance implies this tensor will transform by
\begin{align}
    &R^{i_1}{}_{j_1}\cdots R^{i_M}{}_{j_M} g^{j_1,\ldots,j_M}(R^{-1}x) \notag \\
    &\qquad = \int k^{i_1,\ldots,i_M,i'_1,\ldots,i'_N}(x-x')   R^{i'_1}{}_{j'_1}\cdots R^{i'_N}{}_{j'_N}f^{j'_1,\ldots,j'_N}(R^{-1}x')dx' \ ,
\end{align}
i.e. if we act on a rotated input with our convolution kernel, we expect to get a rotated output.

We make the change of variables $y = R^{-1}x, y' = R^{-1}x'$ to give
\begin{align}
    &R^{i_1}{}_{j_1}\cdots R^{i_M}{}_{j_M} g^{j_1,\ldots,j_M}(y) \notag \\
    &\qquad = \int k^{i_1,\ldots,i_M,i'_1,\ldots,i'_N}(R(y-y'))   R^{i'_1}{}_{j'_1}\cdots R^{i'_N}{}_{j'_N}f^{j'_1,\ldots,j'_N}(y')dy' \ .
\end{align}

Multiplying both sides on the left by $(R^T)^{k_l}{}_{i_l}$ for each $l\in\{1,\ldots,M\}$: since $(R^T)^{k_l}{}_{i_l}\,R^{i_l}{}_{j_l} = (R^TR)^{k_l}{}_{j_l} = \delta^{k_l}{}_{j_l}$, contracting $M$ copies of $R^TR=I$ isolates $g^{k_1,\ldots,k_M}$:
\begin{align}
\label{eq:match2}
g^{k_1,\ldots,k_M}(x)
&= \int (R^T)^{k_1}{}_{i_1}\cdots (R^T)^{k_M}{}_{i_M}\,
    k^{i_1,\ldots,i_M,i'_1,\ldots,i'_N}(R(y-y')) \notag \\
&\qquad {}\times
    R^{i'_1}{}_{j'_1}\cdots R^{i'_N}{}_{j'_N}
    f^{j'_1,\ldots,j'_N}(y') \,dy'.
\end{align}
For this equation to hold for all input fields $f$, the integrands must match pointwise. Matching \eqref{eq:match1} and \eqref{eq:match2} and replacing $y-y'$ with $y$ gives the requirement:
\begin{align}
    k^{k_1,\ldots,k_M,j'_1,\ldots,j'_N}(y) = (R^T)^{k_1}{}_{i_1}\cdots (R^T)^{k_M}{}_{i_M}\; R^{i'_1}{}_{j'_1}\cdots R^{i'_N}{}_{j'_N}\; k^{i_1,\ldots,i_M,i'_1,\ldots,i'_N}(Ry) \ .
\end{align}
Since all quantities are real numbers they may be freely reordered.  There are $M+N=r$ copies of $R$ in total: $M$ transposed copies $(R^T)^{k_l}{}_{i_l}$ acting on the output indices, and $N$ standard copies $R^{i'_l}{}_{j'_l}$ acting on the input indices.

This same derivation applies for any $M$ and $N$ (treating multiplication over the empty set as the identity) and yields exactly the same transformation law as long as $M+N=r$.  The law therefore depends only on the total rank $r$, not on how the rank is split between input and output.  Therefore without loss of generality, the proof in Appendix \ref{app:moment_all} takes $N=r$ and $M=0$, i.e. a mapping from a rank-$r$ tensor-valued function to a scalar-valued function; the result extends immediately to any $(M,N)$ with $M+N=r$.

\section{Proof that every equivariant kernel is a moment kernel.}
\label{app:moment_all}

We restate Theorem~\ref{thm:completeness} as reference. Let $d \geq 2$ and let $p, q \geq 0$. Every linear convolution map from rank-$p$ tensor fields to rank-$q$ tensor fields over $\mathbb{R}^d$ that is equivariant with respect to all orthogonal transformations in $d$ dimensions $R \in O(d)$, i.e., every kernel $k$ of rank $r = p + q$ satisfying \eqref{eq:kernel_general}, can be represented in the moment kernel form:
\begin{align}
k^{i_1,\ldots,i_r}(x) = \sum_\sigma f_\sigma(|x|)\;\delta^{i_{\sigma_{1,a}}i_{\sigma_{1,b}}}\cdots\delta^{i_{\sigma_{s,a}}i_{\sigma_{s,b}}}\; x^{i_{\sigma^C_1}}\cdots x^{i_{\sigma^C_{r-2s}}}
\end{align}
where the sum is over all \emph{signatures} $\sigma$, sets of non-overlapping index pairs drawn from $\{1,\ldots,r\}$, and each $f_\sigma:\mathbb{R}_{\geq 0}\to\mathbb{R}$ is a learnable radial function. 
Here we describe a given $\sigma = \{\{\sigma_{1,a},\sigma_{1,b}\},\ldots,\{\sigma_{s,a},\sigma_{s,b}\}\}$ for $s = |\sigma|$ total index pairs. We denote its complement $\sigma^C = \{1,\ldots,r\}\setminus \cup \sigma = \{\sigma^C_{1},\ldots,\sigma^C_{r-2s}\}$ for $|\sigma^C|=r-2s$ unpaired indices. Each index pair $\{a,b\} \in \sigma $ contributes a Kronecker delta $\delta^{i_a i_b}$.  Each remaining unpaired index $c \in \sigma^C$ is filled by position components $x^{i_c}$. 

The representation need not be unique when these tensors are linearly dependent, but the span is complete. Here we give a proof in the spirit of the main text. The proof works by first looking at the kernel on the positive $e_1$ axis and asking which tensor components can remain unchanged under every rotation and reflection that leaves this axis fixed. We begin with matrices, where the argument is easy to see ($r=2$), and then repeat the same coefficient constraints for arbitrary rank.

\subsection{Values on one axis determine values everywhere}

Fix a radius $t>0$ and write $x_0=t e_1$, where $e_1$ is the first standard basis vector. Note that we could choose any axis, but work with $e_1$ here without loss of generality.
Note we consider the $t=0$ case separately. If $k$ is equivariant and $R e_1 = n$ for some $R\in SO(d)$ and unit vector $n$, then
\begin{align}
    k^{i_1,\ldots,i_r}(tn)
    =
    k^{i_1,\ldots,i_r}(Rx_0)
    =
    R^{i_1}{}_{j_1}\cdots R^{i_r}{}_{j_r}
    k^{j_1,\ldots,j_r}(x_0).
\end{align}
Thus the value of the kernel on the sphere of radius $t$ is determined by its value at $t e_1$.

There is one consistency condition. If a transformation $h\in O(d)$ fixes $e_1$ (i.e. $h e_1 = e_1$), then it fixes $x_0$, so equivariance requires
\begin{align}
\label{eq:axis_invariance}
    k^{i_1,\ldots,i_r}(x_0)
    =
    k^{i_1,\ldots,i_r}(hx_0)
    =
    h^{i_1}{}_{j_1}\cdots h^{i_r}{}_{j_r}
    k^{j_1,\ldots,j_r}(x_0).
\end{align}
In words: the tensor $k(t e_1)$ must be invariant under all rotations and reflections of the  coordinates perpendicular to $e_1$. We now characterize the kernels with this property.

\subsection{The matrix case}

Let $r=2$, so $k(t e_1)$ is a matrix for all $t>0$. Write it in the standard basis as
\begin{align}
    k(t e_1)=\sum_{a=1}^d\sum_{b=1}^d c^{ab}(t)\,e_a e_b^T .
\end{align}
We will determine which coefficients $c^{ab}(t)$ can be nonzero.

\paragraph{Reflections remove mixed axis-perpendicular terms.}
For each $m>1$, let $F_m$ be the reflection that flips $e_m$ and leaves all other coordinate axes fixed:
\begin{align}
    F_m e_m=-e_m,\qquad F_m e_a=e_a\quad (a\neq m).
\end{align}
Consider $m>1$, since $F_m e_1=e_1$, \eqref{eq:axis_invariance} gives for a given $a,b$
\begin{align}
    c^{ab}(t)\,e_a e_b^T
    =
    c^{ab}(t)\,(F_m e_a)(F_m e_b)^T
    =
    (-1)^{\delta_{am}+\delta_{bm}} c^{ab}(t)\,e_a e_b^T
\end{align}
for each basis term after comparing coefficients. Therefore
\begin{align}
\label{eq:matrix_sign_constraint}
    c^{ab}(t)=(-1)^{\delta_{am}+\delta_{bm}}c^{ab}(t).
\end{align}
If exactly one of $a,b$ equals $m$, this equation says $c^{ab}(t)=-c^{ab}(t)$, hence $c^{ab}(t)=0$. Since this holds for every $m>1$, the only coefficients that can survive are $c^{11}(t)$ and the perpendicular diagonal coefficients $c^{mm}(t)$ for $m>1$.

\paragraph{Rotations make all perpendicular diagonal coefficients equal.}
Now choose two perpendicular axes $m,n>1$ and let $Q_{mn}$ be a 90$^\circ$ rotation in their coordinate plane:
\begin{align}
    Q_{mn}e_m=e_n,\qquad Q_{mn}e_n=-e_m,\qquad Q_{mn}e_a=e_a\quad (a\notin\{m,n\}).
\end{align}
Again $Q_{mn}e_1=e_1$, so \eqref{eq:axis_invariance} gives
\begin{align}
    c^{mm}(t)e_m e_m^T+c^{nn}(t)e_n e_n^T
    =
    c^{mm}(t)e_n e_n^T+c^{nn}(t)e_m e_m^T .
\end{align}
Comparing coefficients yields $c^{mm}(t)=c^{nn}(t)$. Therefore all perpendicular diagonal coefficients are equal; call their common value $b(t)$. The most general matrix satisfying \eqref{eq:axis_invariance} is
\begin{align}
    k(t e_1)
    = a(t)e_1e_1^T + b(t)\sum_{m=2}^d e_m e_m^T
    = a(t)e_1e_1^T + b(t)(I-e_1e_1^T).
\end{align}
Equivalently, after combining coefficients to allow the sum to range from 1 to $d$,
\begin{align}
\label{eq:matrix_axis_form}
    k(t e_1)=\alpha(t)e_1e_1^T+\beta(t)I.
\end{align}

Note that for $d=2$ there is only one perpendicular diagonal coefficient, and so this argument is not necessary. i.e. the argument shows that if there is more than one perpendicular diagonal coefficient, then they all must be equal.

\paragraph{Moving off the axis.}
For an arbitrary nonzero $x=t n$, choose $R\in O(d)$ with $R e_1=n=x/|x|$. Using equivariance and \eqref{eq:matrix_axis_form},
\begin{align}
    k(x)
    &= R k(t e_1)R^T\\
    &= \alpha(t)(Re_1)(Re_1)^T+\beta(t)I\\
    &= \alpha(|x|)\frac{xx^T}{|x|^2}+\beta(|x|)I.
\end{align}
Absorbing the factor $|x|^{-2}$ into the radial coefficient gives
\begin{align}
    k^{ij}(x)=f_{22}(|x|)x^ix^j+f_{21}(|x|)\delta^{ij},
\end{align}
which is the rank-2 moment kernel.

\subsection{A coefficient lemma for arbitrary rank}

The matrix proof used only two elementary facts: reflections force perpendicular directions to occur an even number of times, and rotations prevent the choice of a preferred perpendicular direction. We now state the corresponding rank-$r$ version explicitly.

\paragraph{Lemma.}
Let $T$ be a rank $r$ tensor in $d\geq 2$ dimensions such that
\begin{align}
\label{eq:T_axis_invariance}
    T^{i_1,\ldots,i_r}
    =
    h^{i_1}{}_{j_1}\cdots h^{i_r}{}_{j_r}
    T^{j_1,\ldots,j_r}
\end{align}
for every orthogonal transformation $h$ with $h e_1=e_1$, i.e. $T$ is invariant to such transformations $h$.  Then $T$ is a linear combination of tensors formed by placing 
 Kronecker deltas in paired slots $\{\{a_1,b_1\},\ldots,\{a_s,b_s\}\}$, and 
$e_1$ in unpaired slots $\{c_1,\ldots,c_{r-2s}\}$:
\begin{align}
\label{eq:T_axis_span}
    \delta^{i_{a_1}i_{b_1}}\cdots\delta^{i_{a_s}i_{b_s}}
     e_1^{i_{c_1}}\cdots e_1^{i_{c_{r-2s}}}.
\end{align}
Note that, in this lemma, we consider a single tensor, not a tensor field.

\paragraph{Proof of the lemma.}
Write $T$ in a standard basis with coefficients $c$ as
\begin{align}
    T=\sum_{j_1,\ldots,j_r=1}^d
    c^{j_1\ldots j_r}\,
    e_{j_1}\otimes\cdots\otimes e_{j_r}.
\end{align}
We refer to each term in the tensor product as a ``slot''.
For $m>1$, apply the reflection $F_m$ used above. Comparing the coefficient of each basis tensor gives
\begin{align}
\label{eq:rankr_sign_constraint}
    c^{j_1\ldots j_r}
    =
    (-1)^{N_m(j_1,\ldots,j_r)}
    c^{j_1\ldots j_r},
\end{align}
where $N_m(j_1,\ldots,j_r)$ is the number of slots where $j=m$. Hence a coefficient can be nonzero only if every perpendicular label $m>1$ occurs an even number of times.

This proves, in particular, that perpendicular slots cannot occur singly or in odd groups. When two perpendicular slots are present, the same rotation argument as in the matrix case shows that we must have an equal coefficient for all directions in these perpendicular slots.  They are then combined into a single basis element by summing over all perpendicular basis directions, giving the perpendicular projection matrix:
\begin{align}
\label{eq:sumfrom2}
    P^{ij}=\sum_{m=2}^d e_m^i e_m^j
    =\delta^{ij}-e_1^i e_1^j.
\end{align}
Thus a pair of perpendicular slots contributes $P$.

Our axis aligned rotation argument implies that 4 or more perpendicular slots may also share a single coefficient, but the resulting tensor is not invariant under arbitrary rotations in the perpendicular plane. For example, for $L$ slots consider
\begin{align}
    U_L=\sum_{m=2}^d e_m^{\otimes L}.
\end{align}
Rotate in the $(e_2,e_3)$ plane by angle $\theta$:
\begin{align}
    e_2\mapsto \cos\theta\,e_2+\sin\theta\,e_3,\qquad
    e_3\mapsto -\sin\theta\,e_2+\cos\theta\,e_3.
\end{align}
In the rotated tensor, the coefficient of
$e_2\otimes\cdots\otimes e_2\otimes e_3$ is
\begin{align}
\label{eq:larger_tuple_rotation}
    \cos^{L-1}\theta\,\sin\theta
    -
    \sin^{L-1}\theta\,\cos\theta.
\end{align}
For $L=2$ this coefficient is identically zero, which is why a pair gives an invariant matrix. For even $L>2$, it is not identically zero, so such a larger block cannot be invariant. Therefore perpendicular slots must be assembled from pairwise projection matrices rather than larger axis-aligned blocks.

Repeating this pairing argument for all possible choices of slots shows that the equivariant tensors are spanned by products of perpendicular projection matrices $P$ on paired slots and $e_1$ factors on the remaining slots. Finally, expanding the sum  of \eqref{eq:sumfrom2} to range from 1 to $d$ rewrites the same span using only Kronecker deltas and $e_1$ factors,
\begin{align}
    P^{ij}=\delta^{ij}-e_1^i e_1^j \ .
\end{align}
Repeating the argument for arbitrary choices of other paired slots gives \eqref{eq:T_axis_span}. This proves the lemma.

\subsection{Completing the rank-$r$ proof}

Apply the lemma to $T=k(t e_1)$. For every $t>0$,
\begin{align}
\label{eq:axis_form_elementary}
k^{i_1,\ldots,i_r}(t e_1)
 =
 \sum_\sigma a_\sigma(t)\;
\delta^{i_{\sigma_{1,a}}i_{\sigma_{1,b}}}\cdots\delta^{i_{\sigma_{s,a}}i_{\sigma_{s,b}}}\;
e_1^{i_{\sigma^C_{1}}}\cdots e_1^{i_{\sigma^C_{r-2s}}},
\end{align}
where $\sigma$ ranges over all choices of non-overlapping index pairs.

Now let $x\neq 0$, set $t=|x|$, and choose $R\in O(d)$ with $Re_1=x/t$. Equivariance gives
\begin{align}
k^{i_1,\ldots,i_r}(x)
&= R^{i_1}{}_{j_1}\cdots R^{i_r}{}_{j_r}
   k^{j_1,\ldots,j_r}(t e_1)\\
&=
\sum_\sigma a_\sigma(t)\;
\delta^{i_{\sigma_{1,a}}i_{\sigma_{1,b}}}\cdots\delta^{i_{\sigma_{s,a}}i_{\sigma_{s,b}}}\;
(Re_1)^{i_{\sigma^C_1}}\cdots (Re_1)^{i_{\sigma^C_{r-2s}}}.
\end{align}
The Kronecker deltas are unchanged by $R$, and $Re_1=x/|x|$. Hence
\begin{align}
k^{i_1,\ldots,i_r}(x)
&=
\sum_\sigma a_\sigma(|x|)\,|x|^{-(r-2s)}
\delta_{i_{a_1}i_{b_1}}\cdots\delta_{i_{a_s}i_{b_s}}\;
x^{i_{c_1}}\cdots x^{i_{c_{r-2s}}}.
\end{align}
Absorbing the power of $|x|$ into the radial coefficient gives the moment kernel form.

\subsection{The origin}

At $x=0$, equivariance requires
\begin{align}
    k^{i_1,\ldots,i_r}(0)
    =
    R^{i_1}{}_{j_1}\cdots R^{i_r}{}_{j_r}
    k^{j_1,\ldots,j_r}(0)
    \qquad\text{for every }R\in O(d).
\end{align}
Thus $k(0)$ is an $O(d)$-invariant tensor.  The same reflection and rotation arguments above, now with no distinguished $e_1$ direction, imply that only full pairings of all indices with Kronecker deltas can remain. Hence $k(0)=0$ for odd $r$, while for even $r$ it is spanned by terms of the form
\begin{align}
    \delta^{i_{a_1}i_{b_1}}\cdots\delta^{i_{a_{r/2}}i_{b_{r/2}}}.
\end{align}
This is the $x=0$ specialization of the moment kernel family.
In the discrete convolutional kernels used in our experiments, the value at the origin is learned directly through these same admissible tensor structures.

\subsection{Sufficiency}
Finally, every tensor in the moment kernel family is equivariant. Radial functions are invariant because $|Rx|=|x|$; position components transform as $(Rx)^i=R^i{}_j x^j$; and the Kronecker delta is invariant because
\begin{align}
    R^i{}_{k}R^j{}_{l}\delta^{kl}=(RR^T)^{ij}=\delta^{ij}.
\end{align}
Products of these factors therefore transform with one copy of $R$ for each tensor index, exactly as required by \eqref{eq:kernel_general}. This proves both necessity and sufficiency, and completes the proof of Theorem~\ref{thm:completeness}.

\section{SRE-Conv configurations on DermaMNIST}
\label{sec:sre-conv-ablations}

The SRE-Conv kernel of \citet{du2024sreconv} is parameterized as a spatially-averaged filter and is exactly $D_4$-equivariant on the integer grid. The published architecture is a 3.99M-parameter SRE-ResNet18 trained at $32\times32$ input resolution. To compare SRE-Conv against moment kernels at the parameter scale used elsewhere in this paper, we substitute the SRE-Conv kernel into our 8-layer 210k backbone. Because the 8-layer backbone halves spatial resolution four times, the choice of input size and downsampling layer interacts with $D_4$-equivariance: stride-2 average pooling on odd-sized feature maps breaks $D_4$ on the integer grid even when the kernel itself is exact. Table~\ref{tab:sre-conv-ablations} reports three configurations that resolve this interaction in different ways. The first uses our parity-aware \texttt{mk.Downsample} (the same downsampling layer used for moment kernel rows in Table~\ref{tab:classification}) on native $28\times28$ inputs; the second matches the published SRE-Conv recipe by resizing to $32\times32$ and using standard \texttt{AvgPool2d(2,2)}; the third re-evaluates the published SRE-ResNet18 architecture under our worst-case protocol, which also resized to $32\times32$. All three configurations achieve WC drop $= 0$ to four decimal places, confirming that the SRE-Conv kernel itself is exact-$D_4$ regardless of integration choice. At the 210k parameter scale, SRE-Conv reaches 69.9--71.7\% accuracy across configurations and augmentation conditions, well below moment kernels at the larger parameter budget (75.4--76.6\%, Table~\ref{tab:classification}). The published SRE-ResNet18 reaches 76.9\% accuracy under our 8-orientation worst-case protocol (the original work \citep{du2024sreconv} reports 75.7\% on DermaMNIST, and evaluates rotations and reflections on separate test conditions, so it does not report a worst-case metric joint over rotations and reflections), but with $19\times$ more parameters (3.99M vs.\ 210k); at parameter parity, the moment-kernel construction provides a clear accuracy advantage.

\begin{table}[t]
  \centering
  \caption{SRE-Conv configurations on DermaMNIST under our 8-orientation worst-case protocol. All three configurations are exactly $D_4$-equivariant on the integer grid (WC drop $= 0$ to four decimal places). \textit{Top block:} 8-layer 210k parameter-matched backbone with parity-aware \texttt{mk.Downsample} on native $28\times28$ inputs, where the parity-aware downsampling subsamples on odd dimensions and averages pairs on even dimensions. \textit{Middle block:} same 8-layer 210k backbone with standard \texttt{AvgPool2d(2,2)} and inputs resized to $32\times32$, ensuring all intermediate feature sizes are even. This matches the published SRE-Conv default training resolution. \textit{Bottom block:} published SRE-ResNet18 architecture from \citet{du2024sreconv} (3.99M parameters) trained with the published recipe and re-evaluated under our protocol; the ``default'' aug refers to the published training augmentation (\texttt{moco-aug} + \texttt{translation}, no rotation aug).}
  \label{tab:sre-conv-ablations}
  \small
  \begin{tabular}{llrrrrr}
    \toprule
    Configuration & Aug & Params & Acc & WC Acc & AUC & WC AUC \\
    \midrule
    \multicolumn{7}{l}{\textit{8-layer 210k backbone, $28\times28$ input +} \texttt{mk.Downsample}} \\
    SRE-Conv kernel & none       & 210k & 69.9\% & 69.9\% & 0.880 & 0.880 \\
    SRE-Conv kernel & discrete   & 210k & 70.5\% & 70.5\% & 0.896 & 0.896 \\
    SRE-Conv kernel & continuous & 210k & 71.0\% & 71.0\% & 0.891 & 0.891 \\
    \midrule
    \multicolumn{7}{l}{\textit{8-layer 210k backbone, $32\times32$ resize + } \texttt{AvgPool2d(2,2)} (matches published recipe)} \\
    SRE-Conv kernel & none       & 210k & 71.7\% & 71.7\% & 0.892 & 0.892 \\
    SRE-Conv kernel & discrete   & 210k & 71.1\% & 71.1\% & 0.897 & 0.897 \\
    SRE-Conv kernel & continuous & 210k & 70.8\% & 70.8\% & 0.899 & 0.899 \\
    \midrule
    \multicolumn{7}{l}{\textit{Published SRE-ResNet18 architecture} \citep{du2024sreconv}} \\
    SRE-ResNet18 & default    & 3{,}995k & 76.9\% & 76.9\% & 0.921 & 0.921 \\
    \bottomrule
  \end{tabular}
\end{table}

\section{Tensor rank ablation on DermaMNIST}
\label{sec:rank-ablation}

To assess the contribution of each tensor rank, we compare three variants under continuous augmentation: scalars only (rank 0, 174k parameters), scalars and vectors (rank 2, 196k), and the full model (rank 4, 174k). All variants achieve WC drop $= 0\%$, confirming equivariance is preserved at every rank. Accuracy and AUC increase monotonically (Table~\ref{tab:rank-ablation}): 67.6\%/0.854 (rank 0) $\to$ 75.5\%/0.921 (rank 2) $\to$ 76.6\%/0.933 (rank 4). The largest gain comes from adding vector channels (+7.9\% accuracy), with a further improvement from matrix channels (+1.1\%), motivating the full rank-4 design used in our main results.

\begin{table}[h]
  \centering
  \caption{Tensor rank ablation on DermaMNIST. All variants trained under continuous augmentation for 1{,}000 epochs and evaluated under the 8-orientation worst-case protocol. Accuracy and AUC increase monotonically with rank; equivariance is preserved at every rank.}
  \label{tab:rank-ablation}
  \small
  \begin{tabular}{llrrrr}
    \toprule
    Variant & Aug & Params & Acc & WC Acc & AUC \\
    \midrule
    Rank 0 — Scalars only (ours)       & continuous & 174k & 67.6\%          & 67.6\%          & 0.854 \\
    Rank 2 — Scalars + vectors (ours)  & continuous & 196k & 75.5\%          & 75.5\%          & 0.921 \\
    Rank 4 — Moment kernels (ours)     & continuous & 174k & \textbf{76.6\%} & \textbf{76.6\%} & \textbf{0.933} \\
    \bottomrule
  \end{tabular}
\end{table}

\section{Generalization to BloodMNIST}
\label{sec:bloodmnist}
We replicate the DermaMNIST classification experiment on BloodMNIST, an 8-class peripheral blood cell classification benchmark from MedMNIST \citep{yang2023medmnist}, using identical architectures and the same 8-orientation worst-case evaluation protocol. Standard accuracy on BloodMNIST is high for all methods, and the standard CNN with continuous augmentation reaches the top standard accuracy (96.6\%, vs.\ 96.2\% for moment kernels). The methods separate on WC accuracy (Table~\ref{tab:bloodmnist}): moment kernels achieve the highest WC accuracy (96.2\%) with zero WC drop without any rotation augmentation, ESCNN and SRE-Conv also achieve zero WC drop, with lower WC accuracies of 94.9\% and 93.5\%, respectively. In contrast, the standard CNN loses 3.7\% WC accuracy even with continuous augmentation and 17.1\% without augmentation. The WC protocol exposes non-equivariance at specific 90$^\circ +$ reflection combinations. The consistent pattern across two independent biomedical datasets supports the generality of the findings: equivariance by construction matches or exceeds augmentation on standard accuracy while guaranteeing zero WC drop.

\begin{table}[t]
  \centering
  \caption{Image classification on BloodMNIST (same evaluation protocol as Table~\ref{tab:classification}).
  All models trained for 1{,}000 epochs without augmentation unless stated.
  Standard CNN and SRE-Conv are sized to exceed the moment kernel's parameter count.}
  \label{tab:bloodmnist}
  \small
  \begin{tabular}{llrrrr}
    \toprule
    Model & Aug & Params & Acc & WC Acc & AUC  \\
    \midrule
    Moment kernels (ours)          & none       & 174k & 96.2\%          & \textbf{96.2\%} & \textbf{0.998} \\
    ESCNN \citep{cesa2022a}        & none       & 185k & 94.9\%          & 94.9\%        & 0.997          \\
    SRE-Conv \citep{du2024sreconv} & none       & 210k & 93.5\%          & 93.5\%          & 0.995         \\
    Standard CNN                   & continuous & 200k & \textbf{96.6\%} & 93.0\%         & \textbf{0.998} \\
    Standard CNN                   & none       & 200k & 92.7\%          & 75.6\%          & 0.993        \\
    \bottomrule
  \end{tabular}
\end{table}

\section{Generalization experiments for 3D affine registration}
\label{sec:registration-aug}

Two experiments isolate equivariance from the training distribution. Experiment 2 trains both methods without rotation augmentation (only scale and translation) and evaluates under the 48-orientation protocol with $\mathrm{SO}(3)$ random affines; Experiment 3 gives the standard CNN full $\mathrm{O}(3)$ training augmentation while the moment kernel keeps its $\mathrm{SO}(3)$ training, with both evaluated under the 48-orientation protocol with $\mathrm{O}(3)$ random affines. Results are reported in Table~\ref{tab:registration-aug}.

\begin{table}[h]
\centering
\caption{Generalization experiments. Experiment 2: both methods trained without rotation augmentation (only scale and translation), then evaluated under the 48-orientation protocol with $\mathrm{SO}(3)$ random affines. Experiment 3: the standard CNN is given full $\mathrm{O}(3)$ training augmentation while the moment kernel keeps its $\mathrm{SO}(3)$ training; both evaluated under the 48-orientation protocol with $\mathrm{O}(3)$ random affines. Metrics are per-element averages (Methods §\ref{sec:registration}).}
\label{tab:registration-aug}
\small
\begin{tabular}{llccccc}
\hline
Exp & Method & Train aug & Eval aug & MSE & WC-MSE & ED2 \\
\hline
\multirow{2}{*}{2} & Standard CNN  & none & $\mathrm{SO}(3)$ & 1.768 & 3.698 & 1.771 \\
                   & \textbf{Moment kernel (ours)} & none & $\mathrm{SO}(3)$ & $\mathbf{0.456}$ & $\mathbf{0.628}$ & $\mathbf{0.097}$ \\
\hline
\multirow{2}{*}{3} & Standard CNN  & $\mathrm{O}(3)$ & $\mathrm{O}(3)$ & 0.0903 & 0.168 & 0.0417 \\
                   & \textbf{Moment kernel (ours)} & $\mathrm{SO}(3)$ & $\mathrm{O}(3)$ & $\mathbf{0.0139}$ & $\mathbf{0.0206}$ & $\mathbf{0.0051}$ \\
\hline
\end{tabular}
\end{table}

\paragraph{Equivariance handles unseen rotations (Experiment 2).}
When trained without rotation augmentation, the moment kernel handles all 48 orientations consistently (ED2 $=0.097$) despite never having seen a rotated input. The standard CNN with the same training collapses (ED2 $=1.77$, MSE $=1.77$). This isolates equivariance as a consistency guarantee independent of accuracy: the equivariant model's predictions are less accurate than under rotation augmentation but remain orientation-consistent; the standard CNN's are both less accurate and inconsistent across orientations.

\paragraph{Augmentation cannot replicate equivariance (Experiment 3).}
Giving the standard CNN full $\mathrm{O}(3)$ augmentation more than halves its MSE (0.090 vs.\ 0.196), but it remains $6.5\times$ worse than the moment kernel on MSE and $8\times$ worse on ED2. This confirms that augmentation can cover the training distribution but cannot replicate the structural consistency equivariance provides. The moment kernel's performance is nearly unchanged between $\mathrm{SO}(3)$ and $\mathrm{O}(3)$ evaluation (MSE $0.0126 \to 0.0139$, ED2 $0.0048 \to 0.0051$): reflections are covered for free by its built-in $\mathrm{O}(3)$ equivariance.

\section{SynthMorph-style encoder architecture}
\label{sec:synthmorph-arch}

The SynthMorph-style encoder rows in Table~\ref{tab:registration} use the affine sub-network architecture from \citet{hoffmann2021synthmorph}, trained from scratch under our supervised affine-regression objective. Note that we are not using the publicly released SynthMorph method. SynthMorph is a pairwise registration network. It takes a moving and a fixed volume as input, outputs a transform, and is trained with image-similarity losses on synthesized label-driven training pairs. Our task is to predict a known random affine applied to a single volume, so the unsupervised pairwise objective is not directly applicable. The baseline keeps the encoder architecture and replaces the training regime with our parameter-MSE objective; this isolates the architectural-choice question (does a different non-equivariant encoder design from the registration literature close the gap?) without conflating it with the synthesis-driven training.

Architecturally, the baseline is an encoder + global average pool + dense head: four stride-2 \texttt{Conv3D} blocks with \texttt{LeakyReLU}(0.2) activations and no batch normalization, followed by global pool over spatial dimensions and a small MLP head ending in a 12-dimensional linear output. The two parameter scales correspond to two channel-width presets:
\begin{itemize}
\item Default: encoder channels $(256, 256, 256, 256)$, dense head $(256, 64) \to 12$, totalling 5{,}399{,}372 parameters. This matches the published SynthMorph affine-encoder configuration.
\item Matched: encoder channels $(12, 24, 48, 96)$, dense head $(64, 64) \to 12$, totalling 174{,}948 parameters. This narrows the channel widths to parameter-match the moment kernel (174{,}384) and the standard CNN (186{,}996).
\end{itemize}
Both variants are trained with $\mathrm{SO}(3)$ augmentation and evaluated on the same 48-orientation protocol as the other methods. As reported in §\ref{sec:registration}, the default variant reaches the same test MSE as the standard CNN at 187k parameters (0.209 vs.\ 0.196) despite the parameter advantage, and the matched variant has test MSE 0.231, worse than the moment kernel (0.013) and slightly worse than the standard CNN at the same parameter scale. We conclude that on this task neither scaling within the non-equivariant family (default) nor adopting a different non-equivariant encoder design (matched) closes the gap to architectural $\mathrm{O}(3)$ equivariance.

\section{ESCNN attempts for 3D affine regression}
\label{sec:escnn-3d-attempts}

We attempted eight ESCNN \citep{cesa2022a} configurations on the 3D affine regression task described in §\ref{sec:registration}. All used the same input/output contract (1-channel volume in, 12-scalar affine out), training data, and training budget (200 epochs, Adam lr=$10^{-4}$, batch size 16). Architectural choices varied along five axes: gspace group, hidden representation type, nonlinearity, batch normalization strategy, and parameter count.

\begin{table}[h]
\centering
\small
\caption{ESCNN configurations attempted on 3D affine regression. Variants using batch normalization (rows A--E and G) plateau at validation MSE 1.6--2.0, two orders of magnitude above moment kernels (0.011) and an order of magnitude above the standard CNN baseline (0.034). Variant F (no BN, $\mathrm{SO}(3)$) is the only configuration that trains end-to-end (val MSE 0.056), but its $\mathrm{O}(3)$ counterpart (variant H) plateaus despite the same BN setting. Variants E, F, G share architecture and parameter count; only the BN choice differs, isolating the BN effect. Variants F and H share BN (none) and irrep mix; only the gspace differs, isolating the rotation-vs-reflection-equivariance effect.}
\label{tab:escnn-attempts}
\resizebox{\textwidth}{!}{
\begin{tabular}{llllllrr}
\toprule
Variant & Group & Hidden reps & Nonlinearity & BN & Params & Best val MSE \\
\midrule
A & \texttt{flipRot3dOnR3} (O(3)) & irrep(1,1) only & NormNonLin & IID & 188\,k & 2.030 \\
B & \texttt{flipRot3dOnR3} (O(3)) & trivial rep+irrep(1,1)+irrep(0,2) (1:1:1) & GatedNonLin & IID & 274\,k & 1.767 \\
C & \texttt{flipRot3dOnR3} (O(3)) & trivial rep+irrep(1,1)+irrep(0,2)  (1:1:1) & GatedNonLin & IID & 337\,k & 1.717 \\
D & \texttt{fullOctaOnR3} ($\mathrm{O}_h$) & regular rep (48-dim) & PointwiseNonLin & IID & 246\,k & 1.839 \\
E & \texttt{rot3dOnR3} (SO(3)) & trivial rep+irrep(1)+irrep(2)  (5:2:1) \cite{rezvani2026} & GatedNonLin & IID & 439\,k & 1.770 \\
F & \texttt{rot3dOnR3} (SO(3)) & trivial rep+irrep(1)+irrep(2) (5:2:1) \cite{rezvani2026} & GatedNonLin & \textbf{none} & 435\,k & \textbf{0.056} \\
G & \texttt{rot3dOnR3} (SO(3)) & trivial rep+irrep(1)+irrep(2) (5:2:1) \cite{rezvani2026} & GatedNonLin & GNorm & 439\,k & 1.586 \\
H & \texttt{flipRot3dOnR3} (O(3)) & trivial rep+irrep(1,1)+irrep(0,2) (5:2:1) \cite{rezvani2026} & GatedNonLin & none & 417\,k & 1.787 \\
\bottomrule
\end{tabular}
}
\end{table}

Variants A--D explore the gspace group (continuous $\mathrm{O}(3)$, discrete $\mathrm{O}_h$) and hidden representation type (irrep(1,1)/vector-only, mixed trivial/scalar+irrep(1,1)+irrep(0,2) at 1:1:1, regular representation), all with default \texttt{IIDBatchNorm3d}. Variants E--H fix the irrep mix to the 5:2:1 ratio of \citet{rezvani2026} and isolate two architectural axes: variants E, F, G share architecture and parameter count and differ only in batch normalization strategy (\texttt{IIDBatchNorm3d}, none, \texttt{GNormBatchNorm} respectively); variants F and H differ only in gspace (\texttt{rot3dOnR3} vs \texttt{flipRot3dOnR3}) and isolate whether the no-BN training fix transfers to architectural reflection equivariance. We additionally tested anti-aliased average pooling (\texttt{PointwiseAvgPoolAntialiased3D}) in place of strided convolution, in case stride-2 subsampling was introducing aliasing artefacts that disrupt equivariance under non-axis-aligned rotations; this produced slower convergence rather than improvement and was reverted.

Across all BN-on variants (A--E, G), training MSE remains within a factor of 1.05 of validation MSE for the duration of training, indicating underfitting rather than overfitting. The model cannot fit the training distribution, regardless of architectural choices or regularization strategy. Variant G (\texttt{GNormBatchNorm}, the equivariance-preserving alternative used in our 2D classification baseline) shows the same plateau as the IIDBN variants, indicating the failure depends on the presence of batch normalization rather than on the per-channel-vs-per-irrep statistics choice. Single-batch overfit tests confirm sufficient capacity (variant C reaches loss $<10^{-3}$ within 500 steps on a fixed mini-batch, comparable to moment kernels), so the failure is in optimization across the data distribution rather than in expressivity. Variant F (no BN, $\mathrm{SO}(3)$) is the only configuration that trains end-to-end (val MSE 0.056). Variant H (no BN, $\mathrm{O}(3)$) descends initially from a large initialization-shock loss but plateaus at val MSE 1.79; the no-BN configuration that allowed training in variant F does not produce comparable convergence in variant H, so we cannot conclude that ``remove BN'' is a general remedy for ESCNN training in 3D. This is consistent with \citet{billot2024equitrack}'s explicit observation that gated nonlinearities cause vanishing gradients in 3D ESCNN networks. We note that both \citet{billot2024equitrack} and \citet{rezvani2026} use ESCNN as a feature extractor coupled to a separate inference component (closed-form solver and non-equivariant decoder, respectively) rather than directly regressing transformation parameters with the equivariant network end-to-end.

\newpage
\newpage
\section*{NeurIPS Paper Checklist}



\begin{enumerate}

\item {\bf Claims}
    \item[] Question: Do the main claims made in the abstract and introduction accurately reflect the paper's contributions and scope?
    \item[] Answer: \answerYes{} 
    \item[] Justification: We claim to implement a new form of rotation and reflection equivariant convolution kernel, and this implementation is described in Moment kernels section 2.3, and 2.4. We claim to prove that all such equivariant kernels take this form, and this is done in Appendix A and B. We claim to implement classification and image registration in this framework, and these are described in the results, with code included in our zip file (for review) and to be linked on github (in the nonanonymized version if the
paper is accepted). We claim that our approach performs better than standard and equivariant alternative neural network architectures, and this is supported in our experiments, presented in the results section.

\item {\bf Limitations}
    \item[] Question: Does the paper discuss the limitations of the work performed by the authors?
    \item[] Answer: \answerYes{} 
    \item[] Justification: The discussion section describes limitations, including uncertainty about optimal parameter initialization and various nonlinearities used.

\item {\bf Theory assumptions and proofs}
    \item[] Question: For each theoretical result, does the paper provide the full set of assumptions and a complete (and correct) proof?
    \item[] Answer: \answerYes{} 
    \item[] Justification: We include a proof of the transformation law for our equivariant kernels in Appendix A, and a proof of the specific form of equivariant kernels in Appendix B.

    \item {\bf Experimental result reproducibility}
    \item[] Question: Does the paper fully disclose all the information needed to reproduce the main experimental results of the paper to the extent that it affects the main claims and/or conclusions of the paper (regardless of whether the code and data are provided or not)?
    \item[] Answer: \answerYes{} 
    \item[] Justification: We provide code to reproduce our architecture and to reproduce the experiments performed. Note that the brain registration experiment includes data that cannot be shared publicly, researchers can apply to the Alzheimer’s Disease Neuroimaging Initiative (ADNI) for access. While all code is included, results cannot be reproduced without authorized data access.

\item {\bf Open access to data and code}
    \item[] Question: Does the paper provide open access to the data and code, with sufficient instructions to faithfully reproduce the main experimental results, as described in supplemental material?
    \item[] Answer: \answerYes{} 
    \item[] Justification: We provide code to reproduce our architecture and to reproduce the experiments performed. Note that the brain registration experiment includes data that cannot be shared publicly, researchers can apply to the Alzheimer’s Disease Neuroimaging Initiative (ADNI) for access. We describe the python environment used with a python version and requirements file.
    

\item {\bf Experimental setting/details}
    \item[] Question: Does the paper specify all the training and test details (e.g., data splits, hyperparameters, how they were chosen, type of optimizer) necessary to understand the results?
    \item[] Answer: \answerYes{} 
    \item[] Justification: The details are described in the methods section, and code to reproduce the experiments is included.

\item {\bf Experiment statistical significance}
    \item[] Question: Does the paper report error bars suitably and correctly defined or other appropriate information about the statistical significance of the experiments?
    \item[] Answer: \answerYes{} 
    \item[] Justification: We do not make claims of statistical significance, but we do quantify variability of our method by repeating our classification experiment with 5 different random seeds, as described in the section ``variability''.

\item {\bf Experiments compute resources}
    \item[] Question: For each experiment, does the paper provide sufficient information on the computer resources (type of compute workers, memory, time of execution) needed to reproduce the experiments?
    \item[] Answer: \answerYes{} 
    \item[] Justification: We describe our compute resources (GPU or CPU) for each experiment and discuss computation time in the discussion section.
    
\item {\bf Code of ethics}
    \item[] Question: Does the research conducted in the paper conform, in every respect, with the NeurIPS Code of Ethics \url{https://neurips.cc/public/EthicsGuidelines}?
    \item[] Answer: \answerYes{} 
    \item[] Justification: We confirm we have reviewed and followed the code of ethics. All brain images used, which were collected and shared by the Alzheimer’s Disease Neuroimaging Initiative, have had their faces and skulls removed to avoid any potential for identification.

\item {\bf Broader impacts}
    \item[] Question: Does the paper discuss both potential positive societal impacts and negative societal impacts of the work performed?
    \item[] Answer: \answerYes{} 
    \item[] Justification: We do not believe there are potential negative societal impacts.  Other potential impacts are described briefly in the discussion.
    
\item {\bf Safeguards}
    \item[] Question: Does the paper describe safeguards that have been put in place for responsible release of data or models that have a high risk for misuse (e.g., pre-trained language models, image generators, or scraped datasets)?
    \item[] Answer: \answerNA{} 
    \item[] Justification: There are no such risks associated with our work.

\item {\bf Licenses for existing assets}
    \item[] Question: Are the creators or original owners of assets (e.g., code, data, models), used in the paper, properly credited and are the license and terms of use explicitly mentioned and properly respected?
    \item[] Answer: \answerYes{} 
    \item[] Justification: The MedMNIST and ADNI datasets we used are properly credited and described.

\item {\bf New assets}
    \item[] Question: Are new assets introduced in the paper well documented and is the documentation provided alongside the assets?
    \item[] Answer: \answerYes{} 
    \item[] Justification: We provide code in a zip file together with appropriate documentation.

\item {\bf Crowdsourcing and research with human subjects}
    \item[] Question: For crowdsourcing experiments and research with human subjects, does the paper include the full text of instructions given to participants and screenshots, if applicable, as well as details about compensation (if any)? 
    \item[] Answer: \answerNA{} 
    \item[] Justification: We perform no crowdsourcing. Our neuroimaging data was originally collected from human subjects for research purposes, and is being used appropriately based on the Alzheimer’s Disease Neuroimaging Initiative guidelines. Analysis of previously collected and deidentified neuroimaging data is generally considered not human subjects research.

\item {\bf Institutional review board (IRB) approvals or equivalent for research with human subjects}
    \item[] Question: Does the paper describe potential risks incurred by study participants, whether such risks were disclosed to the subjects, and whether Institutional Review Board (IRB) approvals (or an equivalent approval/review based on the requirements of your country or institution) were obtained?
    \item[] Answer: \answerYes{} 
    \item[] Justification: No crowdsourcing. Research using images of humans is limited to secondary analysis of anonymized imaging data which is IRB-exempt.

\item {\bf Declaration of LLM usage}
    \item[] Question: Does the paper describe the usage of LLMs if it is an important, original, or non-standard component of the core methods in this research? Note that if the LLM is used only for writing, editing, or formatting purposes and does \emph{not} impact the core methodology, scientific rigor, or originality of the research, declaration is not required.
    \item[] Answer: \answerNA{} 
    \item[] Justification: LLMs were used for writing and editing the paper and for code refactoring and computational-efficiency improvements.


\end{enumerate}

\end{document}